\documentclass[10pt,twocolumn,letterpaper]{article}

\usepackage[pagenumbers]{cvpr} % To force page numbers, e.g. for an arXiv version

\usepackage{multirow}
\usepackage{nicematrix}

\usepackage[accsupp]{axessibility}  % Improves PDF readability for those with disabilities.

\definecolor{cvprblue}{rgb}{0.21,0.49,0.74}
\usepackage[pagebackref,breaklinks,colorlinks,allcolors=cvprblue]{hyperref}

\def\paperID{5981} % *** Enter the Paper ID here
\def\confName{CVPR}
\def\confYear{2026}

\title{Cluster-Wise Spatio-Temporal Masking for Efficient Video-Language Pretraining}

\author{
    Weijun Zhuang$^{1,2}$\quad
    Yuqing Huang$^{1,2}$\quad
    Weikang Meng$^{1,2}$\quad
    Xin Li$^{2*}$\quad
    Ming Liu$^{1}$\quad\\
    Xiaopeng Hong$^{1}$\quad
    Yaowei Wang$^{1,2}$\quad
    Wangmeng Zuo$^{1}$\thanks{Corresponding author.}\\
    $^1$Harbin Institute of Technology \quad
    $^2$Pengcheng Laboratory\\
    \tt\small \{weijunzhuang.hit, domaingreen2, zacharymengwk, xinlihitsz, cswmzuo\}@gmail.com, \\
    \tt\small csmliu@outlook.com, 
hongxiaopeng@hit.edu.cn, wangyw@pcl.ac.cn
}

\begin{document}
\maketitle
\begin{abstract}
Large-scale video-language pretraining enables strong generalization across multimodal tasks but often incurs prohibitive computational costs. Although recent advances in masked visual modeling help mitigate this issue, they still suffer from two fundamental limitations: severe visual information loss under high masking ratios and temporal information leakage caused by inter-frame correlations. To address these challenges, we propose ClusterSTM, a Cluster-Wise Spatio-Temporal Masking strategy for efficient video-language pretraining. ClusterSTM first performs intra-frame clustering to partition visual tokens into multiple semantically independent clusters, then conducts cluster-wise masking by retaining the token with the highest temporal density within each cluster. Our masking strategy ensure that the retained tokens capture holistic video content while exhibit strong temporal correlation. Additionally, we introduce a video-text relevance reconstruction objective that aligns high-level multimodal semantics beyond conventional visual reconstruction. Extensive experiments across multiple benchmarks demonstrate that ClusterSTM achieves superior performance on video-text retrieval, video question answering, and video captioning tasks, establishing a new state-of-the-art among efficient video-language models.
The code is available at \url{https://github.com/illusion-vlm/ClusterSTM}.
\end{abstract}    
\section{Introduction}
\label{sec:intro}

Video-language models~\cite{wang2024internvideo2,bolya2025perception,wang2023all,wang2022internvideo} construct a semantic bridge between visual dynamics and textual descriptions through large-scale pretraining on massive video-text pairs, enabling flexible adaptation across diverse downstream tasks~\cite{ma2022x,jin2024chat,li2023citetracker,zhuang2025spatial,xiao2021next,huang2024rtracker,yang2023vid2seq}. Although such large-scale pretraining endows these models with powerful generalization capabilities, it inevitably entails substantial computational costs. To mitigate this issue, recent studies~\cite{li2023unmasked,wu2025video,fu2023empirical} have investigated masked visual modeling strategies~\cite{he2022masked}, leveraging its high data efficiency and low memory footprint, which has effectively advanced the research on efficient video-language pretraining.

%##########################################################
\begin{figure}
  \centering
  \includegraphics[width=0.98\linewidth]{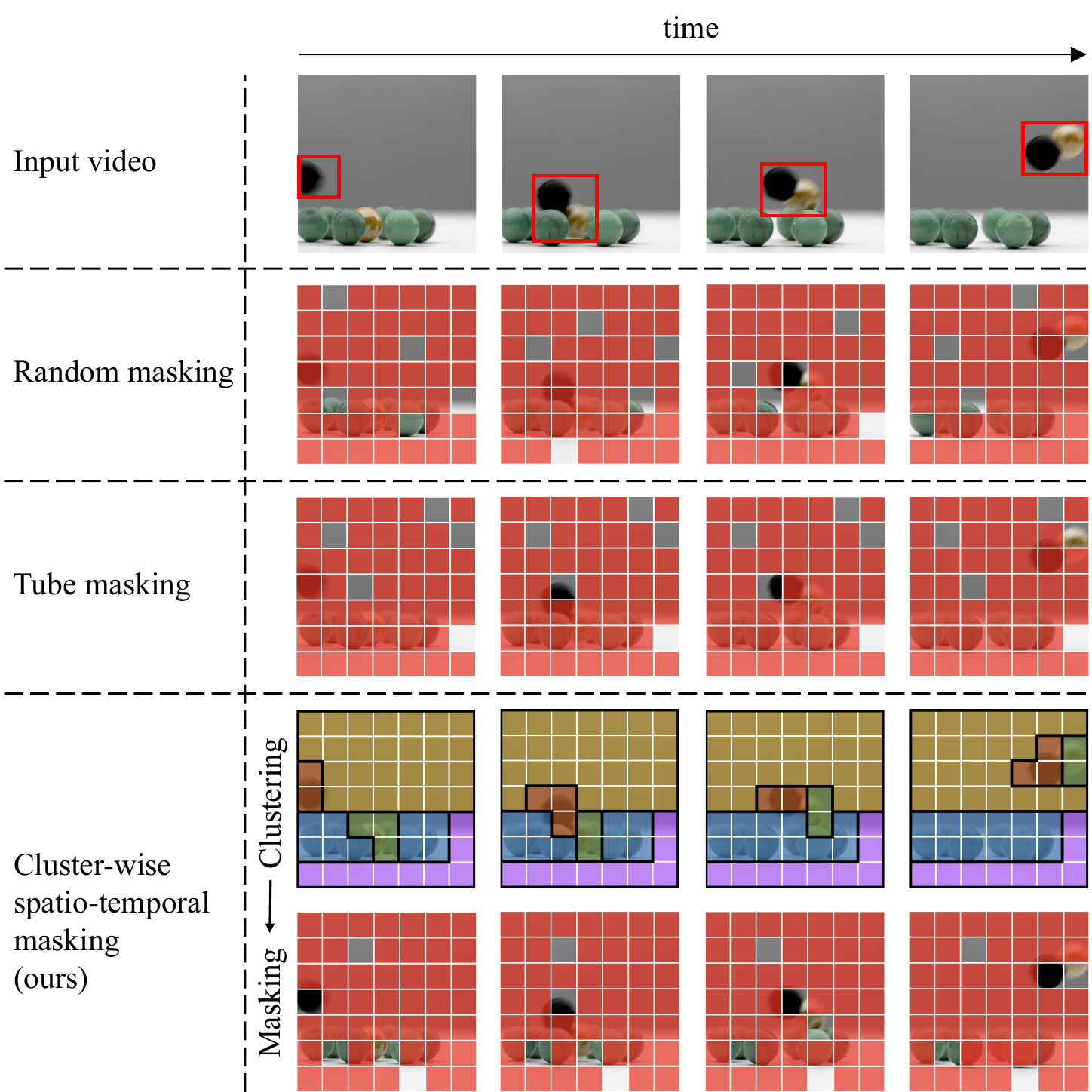}
  \vspace{-1ex}
  \caption{Comparison of different masking strategies. By selecting the token with the highest temporal density within each cluster, the cluster-wise spatio-temporal masking ensures strong temporal correlation among the retained tokens, thereby effectively mitigating the issue of temporal information leakage.}
  \label{fig:temp_leak}
\end{figure}
%##########################################################

The success of masked image/video modeling in masked autoencoders~\cite{he2022masked,tong2022videomae} largely relies on long pretraining schedules (e.g., 2400 epochs on 160K videos) to compensate for the severe loss of visual information caused by high masking ratios (e.g., 0.9). In contrast, large-scale video-language pretraining cannot afford such lengthy training, as models typically converge within only a few epochs (e.g., 10 epochs on 5M video-text pairs). Consequently, a key challenge for efficient video-language pretraining lies in designing an effective masking strategy that selectively retains visual tokens conducive to video-language alignment. To address this limitation, UMT~\cite{li2023unmasked} adopts a semantic masking strategy~\cite{hou2022milan} that selectively retains foreground tokens with high visual semantics for alignment with textual descriptions. However, video-language alignment inherently depends on holistic scene understanding, as accompanying text descriptions usually encompass both foreground and background contents (e.g., \textit{a child is flying a kite on the beach}). Consequently, semantic masking entirely overlooks the contribution of background tokens to vision-language alignment. Meanwhile, STM~\cite{wu2025video} mitigates the problem of severe visual information loss by significantly lowering the masking ratio (e.g., to 0.3) to preserve a large number of visual tokens; nevertheless, this strategy ultimately fails to achieve efficient pretraining.

Another challenge hindering the application of masked video modeling in large-scale video-language pretraining is the problem of information leakage caused by the inherent temporal correlation in video data. Unlike static images, video data contain an additional temporal dimension and exhibit unique properties of temporal correlation. Such temporal correlation may increase the risk of information leakage in the masking and reconstruction pipeline. As illustrated in Figure~\ref{fig:temp_leak}, random masking allows masked tokens to be easily reconstructed by referencing corresponding unmasked tokens from adjacent frames, which undermining the effectiveness of representation learning. The key idea for addressing this problem is to retain tokens that are temporally correlated. Tube masking~\cite{tong2022videomae} achieves this by applying an identical masking map across all frames, thereby preserving tokens at the same spatial locations throughout the video. However, this approach relies on the assumption of minimal motion occurring between adjacent frames, which limits its effectiveness in scenarios characterized by complex motion dynamics.

Building on the above analysis, a masking strategy suitable for efficient video-language pretraining should ensure that the retained tokens not only capture the holistic video content, including both foreground and background information, but also exhibit strong temporal correlation. To this end, we propose a novel masking strategy termed \textbf{ClusterSTM} (\textbf{Cluster}-Wise \textbf{S}patio-\textbf{T}emporal \textbf{M}asking), which effectively addresses the aforementioned challenges while maintaining a high masking ratio. Our masking strategy begins by performing intra-frame clustering over all visual tokens within each video frame, grouping them into multiple semantically independent clusters. Next, for each visual token in a video frame, we compute a temporal density based on its semantic distances to all tokens in adjacent frames. The temporal density of a target token reflects its temporal correlation with tokens in neighboring frames—intuitively, the more semantically similar tokens exist in adjacent frames, the higher the temporal density of the target token. Finally, we perform temporal-density-based cluster-wise masking: within each cluster, we retain the token with the highest temporal density and discard remaining tokens.

%##########################################################
\begin{figure}
  \centering
  \includegraphics[width=0.98\linewidth]{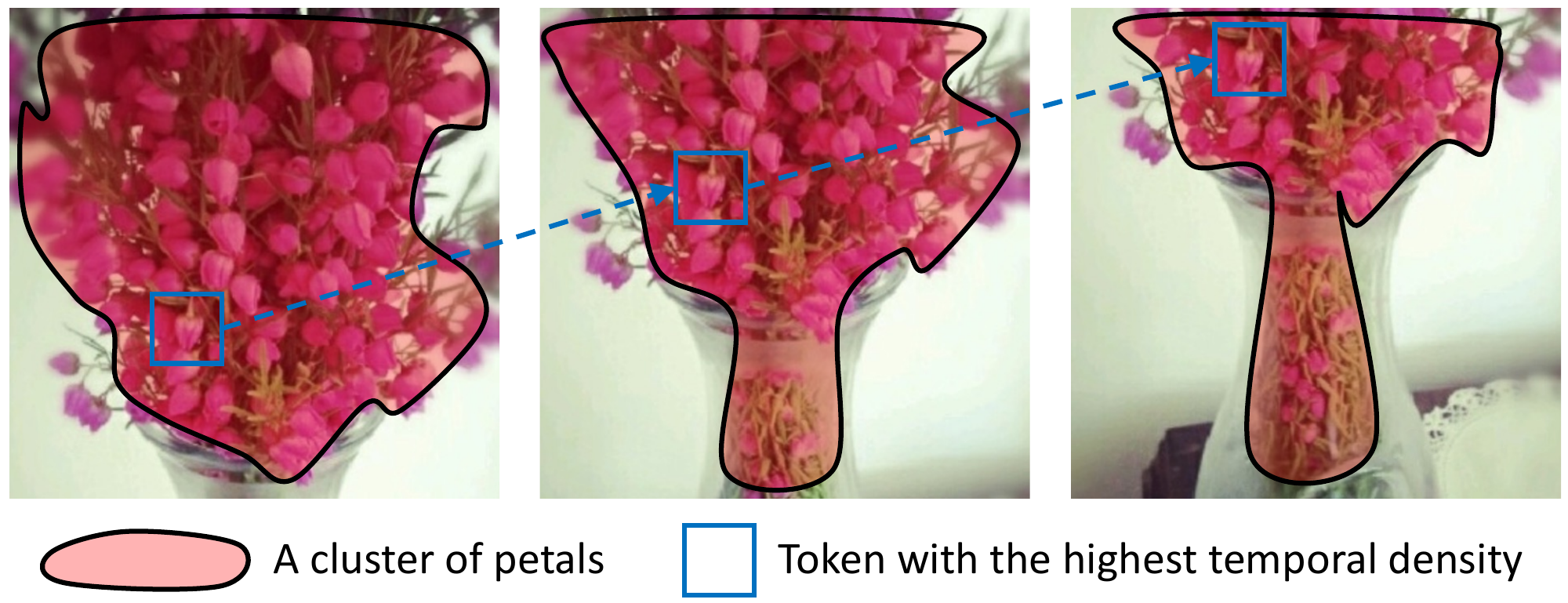}
  \vspace{-1ex}
  \caption{The token with the highest temporal density in a frame typically maintains the highest temporal density in subsequent frames, regardless of its spatial displacement, thereby ensuring that the retained tokens exhibit strong temporal correlation.}
  \label{fig:density}
\end{figure}
%##########################################################

The cluster-wise masking mechanism in ClusterSTM retains one token for each semantically independent cluster, enabling the preserved tokens in a video frame to more comprehensively represent the overall visual content and thereby mitigating visual information loss. Meanwhile, the temporal-density-based token selection strategy ensures that the retained tokens exhibit strong temporal semantic correlation (as illustrated in Figure~\ref{fig:density}), effectively alleviating temporal information leakage. The key contributions of this paper are threefold:

\begin{itemize}
\item[$\bullet$] We present ClusterSTM, a novel masking strategy designed for efficient video-language pretraining. Through intra-frame clustering and subsequent cluster-wise masking, our approach effectively mitigates the severe loss of visual information encountered in prior approaches.
\item[$\bullet$] We introduce a temporal-density-based cluster-wise masking mechanism that computes the temporal semantic density of each visual token and retains token with the highest density within each cluster, thereby ensuring that the preserved tokens exhibit strong temporal correlation.
\item[$\bullet$] We conduct extensive experiments on multiple datasets, demonstrating that ClusterSTM consistently achieves superior performances across various downstream tasks, including text-video retrieval, video question answering, and video captioning.
\end{itemize}

\section{Related Work}
\label{sec:related}

%##########################################################
\begin{figure*}[t]
  \centering
  %\fbox{\rule{0pt}{3in} \rule{0.98\linewidth}{0pt}}
  \includegraphics[width=0.95\linewidth]{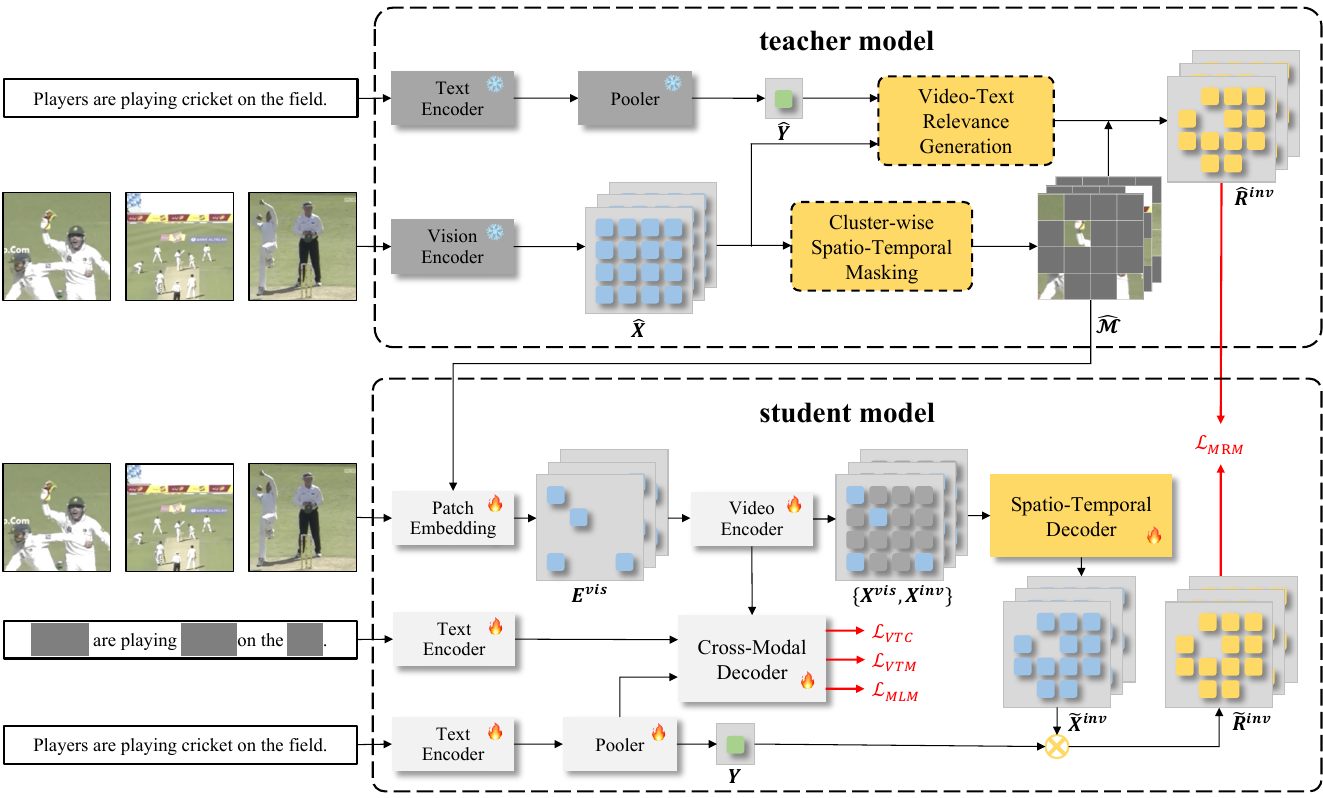}
  \vspace{-1ex}
   \caption{The ClusterSTM pipeline consists of two main components: the Cluster-wise Spatio-Temporal Masking strategy and the Video-Text Relevance Generation process. The Cluster-wise Spatio-Temporal Masking strategy first performs intra-frame clustering, followed by Temporal-Density-based Cluster-wise Masking. In this way, the retained tokens not only comprehensively capture the holistic content of each frame but also exhibit strong temporal semantic consistency. The Video-Text Relevance Generation process then produces fine-grained video-text relevance matrices, which serve as reconstruction targets for MRM loss computation.}
   \label{fig:model}
\end{figure*}
%##########################################################

\noindent\textbf{Video-Language Pretraining.} Video-language pretraining refers to the process of training models that jointly learn representations from both video and text data, with the goal of capturing the complex relationships between visual content and linguistic semantics. Recent advances in this field can be broadly categorized into two families: post-training from image-language models~\cite{wang2022internvideo,jin2024chat,wang2024internvideo2} and training from scratch~\cite{ye2023hitea,wang2024omnivid,bolya2025perception}. The former initializes the model using the weights from well-learned image-language foundation models~\cite{radford2021learning,li2023blip}, whereas the latter starts from random initialization. Despite this difference, both approaches rely on massive video-text pairs for pretraining and therefore face substantial memory and computational requirements. 
Recent works~\cite{MengLLJ025,lei2023revealing,li2023unmasked,wu2025video,meng2025nalaformer,li2023lavender,lin2023smaug} have explored efficient video-language pretraining from different perspectives. Among them, Masked Autoencoder-based methods have demonstrated impressive efficiency gains by masking a large portion of tokens during pretraining. Although these methods have demonstrated impressive efficiency gains, they still suffer from severe visual information loss under high masking ratios as well as temporal information leakage caused by inherent temporal correlations in videos. Our method provides targeted solutions to these limitations by performing cluster-wise spatio-temporal masking.

\noindent\textbf{Masked Visual Modeling.} Following the success of BERT~\cite{devlin2019bert} in natural language processing, a growing line of work has explored masked visual modeling (MVM) for learning visual representations with Vision Transformers~\cite{dosovitskiy2020image}. MAE~\cite{he2022masked} masks a subset of image patches and reconstructs their normalized pixel values, while BEiT~\cite{bao2021beit} learns to predict discrete visual tokens obtained from a pretrained tokenizer. Subsequent approaches have investigated a wide range of target signals for the mask-prediction task, including HOG features~\cite{wei2022masked}, deep features~\cite{baevski2022data2vec,zhou2021ibot}, and frequency components~\cite{liu2023devil,xie2022masked}. For spatio-temporal learning, several methods~\cite{tong2022videomae,feichtenhofer2022masked,wang2022bevt} designs specific MVM strategies tailored to the temporal characteristics of video data. However, despite their effectiveness, these approaches still fail to adequately address the issue of temporal information leakage caused by temporal correlation in video data. Our method tackles this challenge by leveraging temporal density to guide token selection within each cluster to ensure that the retained tokens exhibit strong temporal correlation, thereby effectively mitigates temporal leakage.

\section{Method}

This section introduces the core components of our proposed framework ClusterSTM, designed for efficient video-language pretraining. Specifically, we detail the overall model architecture in Section~\ref{met:overview}, the cluster-wise spatio-temporal masking mechanism in Section~\ref{met:clusterstm}, and the video-text relevance reconstruction in Section~\ref{met:relevance}.

\subsection{Model Overview}
\label{met:overview}

The overall architecture of our model is illustrated in Figure~\ref{fig:model}. We adopt a well-trained vision-language foundation model as the teacher model to generate spatio-temporal masks and video-text relevance matrices. Following UMT~\cite{li2023unmasked}, the student model consists of a video encoder built upon a vanilla ViT~\cite{dosovitskiy2020image} and a text encoder initialized with weights from BERT~\cite{devlin2019bert}.

Given a video-text pair consisting of a video $\mathbf{V}=\{\mathbf{v}_1,...,\mathbf{v}_t,...,\mathbf{v}_T\}$ with $T$ frames and a text sentence $\mathbf{S}$, the teacher model outputs the corresponding video tokens $\mathbf{\widehat{X}}=\{\mathbf{\widehat{x}}_1,...,\mathbf{\widehat{x}}_t,...,\mathbf{\widehat{x}}_T\}\in\mathbb{R}^{T\times{N}\times{D}}$, where $N$ denotes the number of tokens per frame, and the pooled text feature $\mathbf{\widehat{Y}}\in\mathbb{R}^{{1}\times{D}}$. We then apply the cluster-wise spatio-temporal masking strategy to $\mathbf{\widehat{X}}$ to generate the spatio-temporal masks $\mathbf{\widehat{\mathcal{M}}}\in\mathbb{R}^{{T}\times{N}}$ (see Section~\ref{met:clusterstm}). Subsequently, the video-text relevance generation process takes $\mathbf{\widehat{X}}$ and $\mathbf{\widehat{Y}}$ as inputs and produces fine-grained video-text relevance matrices $\mathbf{\widehat{R}}=\{\mathbf{\widehat{r}}_1,...,\mathbf{\widehat{r}}_t,...,\mathbf{\widehat{r}}_T\}\in\mathbb{R}^{{T}\times{N}}$, which serve as the reconstruction targets (see Section~\ref{met:relevance}).

For the student model, we first feed the video $\mathbf{V}$ into a patch embedding module to obtain video embeddings $\mathbf{E}\in\mathbb{R}^{{T}\times{N}\times{D}}$. Guided by the spatio-temporal masks $\mathbf{\widehat{\mathcal{M}}}$ generated by the teacher model, we select the visible video embeddings $\mathbf{E}^{vis}$ and input them into the video encoder to produce the visible video tokens $\mathbf{X}^{vis}$. We then combine the visible and masked (invisible) tokens, denoted as $\{\mathbf{X}^{vis}, \mathbf{X}^{inv}\}$, and feed them into a spatio-temporal decoder, composed of multiple transformer layers similar to those in the video encoder, to reconstruct the full sequence of video tokens $\mathbf{\widetilde{X}}$. For the text sentence $\mathbf{S}$, we pass it through the text encoder to obtain the pooled text feature $\mathbf{Y}$. Finally, we compute the reconstructed video-text relevance matrices $\mathbf{\widetilde{R}}$ by multiplying the reconstructed invisible video tokens $\mathbf{\widetilde{X}}^{inv}$ with the text features $\mathbf{Y}$.

\subsection{Cluster-Wise Spatio-Temporal Masking}
\label{met:clusterstm}

Our masking strategy aims to ensure that the retained visual tokens not only effectively represent the holistic visual content of each frame but also preserve those tokens with strong semantic correlation temporally, therefore effectively addresses the severe visual information loss and temporal information leakage issues encountered when applying masked video modeling to large-scale video-language pretraining. In the following, we detail our proposed cluster-wise spatio-temporal masking method.

\noindent\textbf{Intra-Frame Clustering.} To ensure that the retained visual tokens effectively represent the holistic visual content of each frame, we need to distinguish different semantic regions within the frame. Therefore, we perform intra-frame clustering on each video frame. Specifically, the number of clusters for each frame is defined as $N_c={N}\times{(1-r)}$, where $r$ denotes the masking ratio and $N$ is the number of visual tokens per frame. Given the output visual tokens $\mathbf{\widehat{x}_t}$ at frame $t$ from teacher model, we employ an off-the-shelf clustering algorithm, the Density Peaks Clustering (DPC) algorithm~\cite{rodriguez2014clustering}, to partition $\mathbf{\widehat{x}}_t$ into $N_c$ clusters:

%##########################################################
\begin{equation}
    \mathbb{C}_t=DPC(\mathbf{\widehat{x}}_t,N_c),
\end{equation}
%##########################################################

\noindent\textbf{Temporal-Density-based Cluster-Wise Masking.} Although randomly selecting one token from each semantically independent cluster can effectively preserve the holistic visual content of a video frame, this simple strategy fails to mitigate temporal information leakage. To address this issue, we propose a fine-grained masking mechanism based on temporal density. Specifically, we first define the semantic distance between two vectors as:

%##########################################################
\begin{equation}
    d(\mathbf{u}_i,\mathbf{u}_j)=1-\frac{{\mathbf{u}_i}\cdot{\mathbf{u}_j}}{||\mathbf{u}_i||_2||\mathbf{u}_j||_2},
\end{equation}
%##########################################################

To capture temporal semantic consistency, we refine the local density computation in DPC~\cite{rodriguez2014clustering} and define a temporal density as follows:

%##########################################################
\begin{equation}
    \mathbf{\rho}(\mathbf{\widehat{x}}_{t,n})=\sum_{\substack{i=1 \\ {i}\neq{t}}}^{T}\sum_{j=1}^{N}\exp^{-d(\mathbf{\widehat{x}}_{t,n},\mathbf{\widehat{x}}_{i,j})/d_c},
\end{equation}
%##########################################################
where $\mathbf{\widehat{x}}_{t,n}\in{\mathbb{R}^{1\times{D}}}$ is the $n$-th video token of frame $t$, and $d_c\in\mathbb{R}^+$ is a cutoff distance. Finally, for frame $t$, we retain the token with the highest temporal density within each cluster and discard the remaining tokens, thereby generating the spatio-temporal mask $\mathbf{\widehat{\mathcal{M}}}_t$. 

Although the spatial position of the token with the highest temporal semantic density may shift across subsequent frames, it typically maintains the highest temporal density and thus remains consistently preserved. As a result, our masking strategy ensures that the retained tokens exhibit strong temporal correlation, effectively mitigating the issue of temporal information leakage.

\subsection{Video-Text Relevance Reconstruction}
\label{met:relevance}

Previous video-language pretraining approaches~\cite{fu2023empirical,shu2022masked} have shown that reconstructing high-level spatial features is more advantageous for cross-modal alignment than reconstructing low-level pixels. STM~\cite{wu2025video} further demonstrates the effectiveness of reconstructing even higher-level spatio-temporal representations. However, these methods restrict the reconstruction targets to the visual modality alone, overlooking the collaborative role of the text modality in cross-modal alignment. To address this limitation, we introduce a reconstruction target that operates at a higher semantic level than visual features and inherently carries multimodal properties, termed video-text relevance.

\begin{figure}[t]
  \centering
  \includegraphics[width=0.8\linewidth]{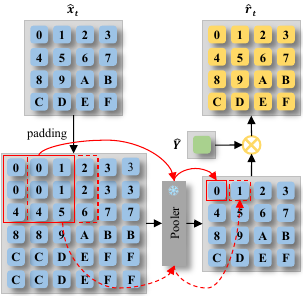}
  \vspace{-1ex}
   \caption{A schematic illustration of the Video-Text Relevance Generation process. The module first aggregates each target token with its neighboring tokens through a pooling operator to obtain an enhanced token. This enhanced token is then multiplied with the text feature to produce a high-quality video-text relevance matrix.}
   \label{fig:rel_rec}
\end{figure}

\noindent\textbf{Video-Text Relevance Generation.} Our process for generating high-quality video-text relevance matrices is illustrated in Figure~\ref{fig:rel_rec}. Since the teacher model is trained with vision-language contrastive learning, it primarily attends to global semantic information. Consequently, directly interacting each visual token with the text feature tends to produce low-quality relevance matrices. To enhance the quality of the relevance representations, we instead aggregate the neighboring tokens surrounding each target token to form a locally enhanced, information-rich token before interacting it with the text feature.

Specifically, we first apply border padding to the visual tokens $\mathbf{\widehat{x}}_t$.
%output by the teacher model. 
Then, we extract local groups of tokens through a sliding window and feed them into the teacher model's pooling module to generate fused visual tokens. Finally, each fused token is multiplied with the text feature to obtain the video-text relevance matrix $\mathbf{\widehat{r}}_t$. We adopt SigLIP~\cite{zhai2023sigmoid} as the teacher model, as its flexible attentive pooling mechanism allows an arbitrary number of visual tokens to be processed, making it highly compatible with the design of our relevance generation process.

\noindent\textbf{Masked Relevance Modeling.} We enhance the model’s ability to learn joint visual-textual representations by reconstructing the video-text relevance. The masked relevance modeling loss uses the teacher-generated relevance matrices $\mathbf{\widehat{R}}^{inv}$ as the reconstruction targets to supervise the reconstruction of $\mathbf{\widetilde{R}}^{inv}$ output by student model:
\begin{equation}
    \mathcal{L}_{MRM}=L2(\mathbf{\widetilde{R}}^{inv},\mathbf{\widehat{R}}^{inv}),
\end{equation}
where $L2(\cdot)$ denotes the L2 distance.

\noindent\textbf{Pretraining Objectives.} In addition to the masked relevance modeling loss, following prior work~\cite{wu2025video,li2023unmasked,fu2023empirical}, we incorporate video-text contrastive learning (VTC) and video-text matching (VTM) loss for cross-modal feature alignment, as well as masked language modeling (MLM) loss to supervise text reconstruction. The overall training objective during pretraining is:
\begin{equation}
    \mathcal{L}=\mathcal{L}_{MRM}+\mathcal{L}_{VTC}+\mathcal{L}_{VTM}+\mathcal{L}_{MLM}.
\end{equation}

\begin{table}
  \caption{Zero-shot text-to-video retrieval Recall@1 results on MSRVTT, DiDeMo, ActivityNet (ANet) and MSVD. \#Pairs denotes the number of pretraining pairs, and \#F indicates the number of video frames sampled during evaluation. The best results are indicated in \textbf{bold}. Models pretrained with massive datasets are noted in \textcolor{gray}{gray}.}
  \label{tab:zs}
  \centering
  \resizebox{0.45\textwidth}{!}{
  \setlength\tabcolsep{2.0pt}
  \begin{NiceTabular}{lrccccc}
  \CodeBefore
    \rowcolors{3}{gray!10}{white}
  \Body
    \toprule
    Method & \#Pairs & \#F & MSRVTT & DiDeMo & ANet & MSVD \\
    \midrule
    Frozen~\cite{bain2021frozen} & 5M & 4 & 18.7 & 20.2 & - & - \\
    VIOLETv2~\cite{fu2023empirical} & 5M & 5 & 21.9 & 24.4 & - & - \\
    Singularity~\cite{lei2023revealing} & 5M & 12 & 28.4 & \textbf{36.9} & 30.8 & - \\
    UMT~\cite{li2023unmasked} & 5M & 4 & 29.6 & 33.4 & 28.3 & 36.2 \\
    STM~\cite{wu2025video} & 5M & 4 & 29.8 & 34.4 & 30.6 & 38.7 \\
    ClusterSTM & 5M & 4 & \textbf{31.2} & 36.5 & \textbf{31.4} & \textbf{40.3} \\
    \midrule
    \multicolumn{6}{l}{\textcolor{gray}{\emph{Models pretrained on more data}}} \\
    \midrule
    \textcolor{gray}{VIOLET}~\cite{fu2021violet} & \textcolor{gray}{138M} & \textcolor{gray}{5} & \textcolor{gray}{25.9} & \textcolor{gray}{23.5} & \textcolor{gray}{-} & \textcolor{gray}{-} \\
    \textcolor{gray}{OmniVL}~\cite{wang2022omnivl} & \textcolor{gray}{17M} & \textcolor{gray}{8} & \textcolor{gray}{34.6} & \textcolor{gray}{33.3} & \textcolor{gray}{-} & \textcolor{gray}{-} \\
    \textcolor{gray}{VINDLU}~\cite{cheng2023vindlu} & \textcolor{gray}{25M} & \textcolor{gray}{4} & \textcolor{gray}{32.0} & \textcolor{gray}{36.9} & \textcolor{gray}{30.9} & \textcolor{gray}{-} \\
    \textcolor{gray}{CLIP4Clip}~\cite{luo2022clip4clip} & \textcolor{gray}{400M} & \textcolor{gray}{12} & \textcolor{gray}{30.6} & \textcolor{gray}{-} & \textcolor{gray}{-} & \textcolor{gray}{36.2} \\
    \textcolor{gray}{InternVideo}~\cite{wang2022internvideo} & \textcolor{gray}{646M} & \textcolor{gray}{8} & \textcolor{gray}{40.7} & \textcolor{gray}{31.5} & \textcolor{gray}{30.7} & \textcolor{gray}{43.4} \\
    \bottomrule
  \end{NiceTabular}
  }
\end{table}

\begin{table*}
  \caption{Text-to-video retrieval on MSRVTT, DiDeMo and AcitivityNet. \#Pairs denotes the number of pretraining pairs. The best results are indicated in \textbf{bold}. Models pretrained with massive datasets are noted in \textcolor{gray}{gray}.}
  \label{tab:t2v}
  \centering
  \resizebox{0.90\textwidth}{!}{
  \setlength\tabcolsep{10pt}
  \begin{NiceTabular}{lrccccccccc}
  \CodeBefore
    \rowcolors{3}{white}{gray!10}
  \Body
    \toprule
    \multirow{2}{*}{Method} & \multirow{2}{*}{\#Pairs} & \multicolumn{3}{c}{MSRVTT} & \multicolumn{3}{c}{DiDeMo} & \multicolumn{3}{c}{ActivityNet} \\
    \cmidrule(lr){3-5}\cmidrule(lr){6-8}\cmidrule(lr){9-11}
    & & R1 & R5 & R10 & R1 & R5 & R10 & R1 & R5 & R10 \\
    \midrule
    ClipBERT~\cite{lei2021less} & 0.2M & 22.0 & 46.8 & 59.9 & 20.4 & 48.0 & 60.8 & 21.3 & 49.0 & 63.5 \\
    Frozen~\cite{bain2021frozen} & 5M & 31.0 & 59.5 & 70.5 & 31.0 & 59.8 & 72.4 & - & - & - \\
    ALPRO~\cite{li2022align} & 5M & 33.9 & 60.7 & 73.2 & 35.9 & 67.5 & 78.8 & - & - & - \\
    BridgeFormer~\cite{ge2022bridging} & 5M & 37.6 & 64.8 & 75.1 & 37.0 & 62.2 & 73.9 & - & - & - \\
    Singularity~\cite{lei2023revealing} & 5M & 36.8 & 65.9 & 75.5 & 47.4 & 75.2 & 84.0 & 43.0 & 70.6 & 81.3 \\
    LAVENDER~\cite{li2023lavender} & 5M & 37.8 & 63.8 & 75.0 & 47.4 & 74.7 & 82.4 & - & - & - \\
    VIOLETv2~\cite{fu2023empirical} & 5M & 37.2 & 64.8 & 75.8 & 47.9 & 76.5 & 84.1 & - & - & - \\
    HiTeA~\cite{ye2023hitea} & 5M & 44.4 & 69.3 & 78.9 & 51.8 & 79.1 & 85.3 & 45.1 & 73.5 & 84.2 \\
    VINDLU~\cite{cheng2023vindlu} & 5M & 43.8 & 70.3 & 79.5 & 54.6 & 81.3 & 89.0 & 51.1 & 79.2 & 88.4 \\
    UMT~\cite{li2023unmasked} & 5M & 46.3 & 72.7 & 82.0 & 54.8 & 83.0 & 89.0 & 52.1 & 80.5 & 89.6 \\
    STM~\cite{wu2025video} & 5M & 48.5 & 74.6 & 83.8 & 56.9 & 84.1 & 89.7 & 53.6 & 82.1 & 90.4 \\
    ClusterSTM & 5M & \textbf{49.7} & \textbf{76.1} & \textbf{85.3} & \textbf{58.5} & \textbf{85.0} & \textbf{90.2} & \textbf{54.9} & \textbf{82.9} & \textbf{90.7} \\
    \midrule
    \multicolumn{6}{l}{\textcolor{gray}{\emph{Models pretrained on more data}}} \\
    \midrule
    \textcolor{gray}{VIOLET}~\cite{fu2021violet} & \textcolor{gray}{138M} & \textcolor{gray}{34.5} & \textcolor{gray}{63.0} & \textcolor{gray}{73.4} & \textcolor{gray}{32.6} & \textcolor{gray}{62.8} & \textcolor{gray}{74.7} & \textcolor{gray}{-} & \textcolor{gray}{-} & \textcolor{gray}{-} \\
    \textcolor{gray}{All-in-one}~\cite{wang2023all} & \textcolor{gray}{138M} & \textcolor{gray}{37.9} & \textcolor{gray}{68.1} & \textcolor{gray}{77.1} & \textcolor{gray}{32.7} & \textcolor{gray}{61.4} & \textcolor{gray}{73.5} & \textcolor{gray}{22.4} & \textcolor{gray}{53.7} & \textcolor{gray}{67.7} \\
    \textcolor{gray}{CLIP4Clip}~\cite{luo2022clip4clip} & \textcolor{gray}{400M} & \textcolor{gray}{42.1} & \textcolor{gray}{71.9} & \textcolor{gray}{81.4} & \textcolor{gray}{43.4} & \textcolor{gray}{70.2} & \textcolor{gray}{80.6} & \textcolor{gray}{40.5} & \textcolor{gray}{72.4} & \textcolor{gray}{-} \\
    \textcolor{gray}{UCOFIA}~\cite{wang2023unified} & \textcolor{gray}{400M} & \textcolor{gray}{49.4} & \textcolor{gray}{72.1} & \textcolor{gray}{-} & \textcolor{gray}{46.5} & \textcolor{gray}{74.8} & \textcolor{gray}{-} & \textcolor{gray}{45.7} & \textcolor{gray}{76.0} & \textcolor{gray}{-} \\
    \textcolor{gray}{UMT-L}~\cite{li2023unmasked} & \textcolor{gray}{25M} & \textcolor{gray}{58.8} & \textcolor{gray}{81.0} & \textcolor{gray}{87.1} & \textcolor{gray}{70.4} & \textcolor{gray}{90.1} & \textcolor{gray}{93.5} & \textcolor{gray}{66.8} & \textcolor{gray}{89.1} & \textcolor{gray}{94.9} \\
    \textcolor{gray}{VAST}~\cite{chen2023vast} & \textcolor{gray}{152M} & \textcolor{gray}{63.9} & \textcolor{gray}{-} & \textcolor{gray}{-} & \textcolor{gray}{72.0} & \textcolor{gray}{-} & \textcolor{gray}{-} & \textcolor{gray}{70.5} & \textcolor{gray}{-} & \textcolor{gray}{-} \\
    \bottomrule
  \end{NiceTabular}
  }
\end{table*}

\vspace{-4ex}

\section{Experiments}

\subsection{Experiment Setup}

\noindent\textbf{Pretraining Datasets.} Following prior methods~\cite{fu2023empirical,li2023unmasked,ye2023hitea,wu2025video}, we pretrain our model on two large-scale datasets, WebVid-2M~\cite{bain2021frozen} and CC3M~\cite{sharma2018conceptual}. WebVid-2M is a web-curated video-text corpus comprising roughly 2.5M video-caption pairs collected from stock footage platforms, whereas CC3M contains 3M web-sourced image-text pairs with diverse, automatically filtered descriptions. The combination of these complementary video and image datasets provide broad and diverse visual-language supervision, enabling the model to jointly benefit from dynamic motion cues and rich static visual semantics.

\noindent\textbf{Implementation Details.} Our model is trained on 8 NVIDIA A100 GPUs using PyTorch. We adopt ViT-B/16~\cite{dosovitskiy2020image} without a class token as video encoder and initialize it with the pretrained weights from UMT~\cite{li2023unmasked}, which is pretrained on Kinetics-710~\cite{li2022uniformerv2}. We initialize the text encoder with weights from BERT-base~\cite{devlin2019bert}.We adopt SigLIP-base~\cite{zhai2023sigmoid} as the teacher model to leverage its flexible attention pooling mechanism to generate the relevance matrix. For the masking strategy, we apply a $75\%$ masking ratio to image tokens and a $90\%$ masking ratio to video tokens. We apply random cluster-wise spatial masking to image tokens and cluster-wise spatio-temporal masking to video tokens. The cutoff distance $d_c$ used in the temporal density computation is set to the top $20\%$ quantiles of the semantic distances. The model is trained on the combined WebVid-2M and CC3M datasets for 10 epochs, with batch size of 1024 for video and 4096 for image, using the AdamW optimizer~\cite{loshchilov2017fixing} with a base learning rate of 1e-4. 

\subsection{Quantitative Results and Comparison}

To comprehensively evaluate our method, we benchmark it against recent state-of-the-art video-language models across three representative video-language tasks, including \textbf{video-text retrieval} (MSRVTT~\cite{xu2016msr}, DiDeMo~\cite{anne2017localizing}, MSVD~\cite{chen2011collecting}, and ActivityNet~\cite{krishna2017dense}), \textbf{video question answering} (TGIF-Action~\cite{jang2017tgif}, TGIF-Transition~\cite{jang2017tgif}, TGIF-Frame~\cite{jang2017tgif}, MSRVTT-MC~\cite{yu2018joint}, MSRVTT-QA~\cite{yu2018joint}, ActivityNet-QA~\cite{yu2019activitynet}, and MSVD-QA~\cite{xu2017video}), and \textbf{video captioning} (MSRVTT~\cite{xu2016msr} and MSVD~\cite{chen2011collecting}).

\noindent\textbf{Zero-Shot Text-to-Video Retrieval.} Table~\ref{tab:zs} reports the zero-shot text-to-video retrieval performance of our method across multiple datasets. Under a comparable pretraining data scale, ClusterSTM achieves remarkable improvements over state-of-the-art models, surpassing them by $\mathbf{1.4\%}$ on MSRVTT, $\mathbf{0.8\%}$ on ActivityNet, and $\mathbf{1.6\%}$ on MSVD. Although ClusterSTM attains performance comparable to Singularity~\cite{lei2023revealing} on DiDeMo, this comparison is skewed by the fact that Singularity samples 12 frames during evaluation, whereas ClusterSTM uses only 4 frames. When the number of campled frames is matched to 4, ClusterSTM outperforms STM~\cite{wu2025video} on DiDeMo by $\mathbf{2.1\%}$. Overall, these results demonstrate the strong generalization ability and efficiency of ClusterSTM across diverse retrieval benchmarks.

\noindent\textbf{Text-to-Video Retrieval.} Table~\ref{tab:t2v} presents the finetuning results for text-to-video retrieval, where ClusterSTM significantly outperforms previous methods pretrained with a comparable data scale. Specifically, ClusterSTM achieves state-of-the-art results with Recall@1 scores of $49.7\%$($\mathbf{+1.2\%}$), $58.5\%$($\mathbf{+1.6\%}$), and $54.9\%$($\mathbf{+1.3\%}$) on MSRVTT, DiDeMo and ActivityNet datasets, respectively. Notably, despite using substantially less pretraining data, ClusterSTM achieves comparable or even superior performance to methods pretrained on massive datasets.

\begin{table*}
  \caption{Video QA results on TGIF, MSRVTT and ActivityNet. The best results are indicated in \textbf{bold}. Models pretrained with massive datasets are noted in \textcolor{gray}{gray}.}
  \label{tab:qa}
  \centering
  \resizebox{0.90\textwidth}{!}{
  \setlength\tabcolsep{10pt}
  \begin{NiceTabular}{lrccccccc}
  \CodeBefore
    \rowcolors{3}{gray!10}{white}
  \Body
    \toprule
    \multirow{2}{*}{Method} & \multirow{2}{*}{\#Pairs} & \multicolumn{3}{c}{TGIF} & \multicolumn{2}{c}{MSRVTT} & \multirow{2}{*}{ANet-QA} & \multirow{2}{*}{MSVD-QA} \\
    \cmidrule(lr){3-5}\cmidrule(lr){6-7}
    & & Act. & Trans & Frame & MC & QA & & \\
    \midrule
    ClipBERT~\cite{lei2021less} & 0.2M & 82.8 & 87.8 & 60.3 & 88.2 & 37.4 & - & - \\
    ALPRO~\cite{li2022align} & 5M & - & - & - & - & 42.1 & - & 45.9 \\
    %Clover~\cite{huang2023clover} & 5M & 94.9 & 98.0 & 71.4 & 95.0 & 43.9 & - \\
    Singularity~\cite{lei2023revealing} & 5M & - & - & - & 92.0 & 42.7 & 41.8 & - \\
    VIOLETv2~\cite{fu2023empirical} & 5M & 94.8 & 99.0 & 72.8 & \textbf{97.6} & 44.5 & - & - \\
    LAVENDER~\cite{li2023lavender} & 5M & 96.6 & 99.1 & 72.2 & 96.6 & 44.2 & - & - \\
    HiTeA~\cite{ye2023hitea} & 5M & 96.8 & 98.8 & 72.5 & 97.2 & 45.4 & 45.1 & - \\
    UMT~\cite{li2023unmasked} & 5M & - & - & - & 95.9 & 44.3 & 43.5 & 49.1 \\
    STM~\cite{wu2025video} & 5M & 96.9 & 99.0 & 73.0 & 96.8 & 45.9 & 44.9 & - \\
    ClusterSTM & 5M & \textbf{96.9} & \textbf{99.1} & \textbf{73.6} & 97.1 & \textbf{46.7} & \textbf{45.5} & \textbf{51.3} \\
    \midrule
    \multicolumn{6}{l}{\textcolor{gray}{\emph{Models pretrained on more data}}} \\
    \midrule
    \textcolor{gray}{JustAsk}~\cite{yang2021just} & \textcolor{gray}{69M} & \textcolor{gray}{-} & \textcolor{gray}{-} & \textcolor{gray}{-} & \textcolor{gray}{-} & \textcolor{gray}{41.5} & \textcolor{gray}{38.9} & \textcolor{gray}{47.5} \\
    \textcolor{gray}{MERLOT}~\cite{zellers2021merlot} & \textcolor{gray}{180M} & \textcolor{gray}{94.0} & \textcolor{gray}{96.2} & \textcolor{gray}{69.5} & \textcolor{gray}{90.9} & \textcolor{gray}{43.1} & \textcolor{gray}{41.4} & \textcolor{gray}{-} \\
    \textcolor{gray}{All-in-one}~\cite{wang2023all} & \textcolor{gray}{283M} & \textcolor{gray}{95.5} & \textcolor{gray}{94.7} & \textcolor{gray}{66.3} & \textcolor{gray}{92.3} & \textcolor{gray}{46.8} & \textcolor{gray}{-} & \textcolor{gray}{47.9} \\
    \textcolor{gray}{MV-GPT}~\cite{seo2022end} & \textcolor{gray}{53M} & \textcolor{gray}{-} & \textcolor{gray}{-} & \textcolor{gray}{-} & \textcolor{gray}{-} & \textcolor{gray}{41.7} & \textcolor{gray}{39.1} & \textcolor{gray}{-} \\
    \textcolor{gray}{VIOLET}~\cite{fu2021violet} & \textcolor{gray}{138M} & \textcolor{gray}{92.5} & \textcolor{gray}{95.7} & \textcolor{gray}{68.9} & \textcolor{gray}{91.9} & \textcolor{gray}{43.9} & \textcolor{gray}{38.9} & \textcolor{gray}{47.9} \\
    \textcolor{gray}{UMT-L}~\cite{li2023unmasked} & \textcolor{gray}{25M} & \textcolor{gray}{-} & \textcolor{gray}{-} & \textcolor{gray}{-} & \textcolor{gray}{97.3} & \textcolor{gray}{47.1} & \textcolor{gray}{47.9} & \textcolor{gray}{55.2} \\
    \bottomrule
  \end{NiceTabular}
  }
\end{table*}

\noindent\textbf{Video Question Answering.} As shown in Table~\ref{tab:qa}, ClusterSTM achieves significant performance improvements in the finetuning results for the video question answering task. Specifically, ClusterSTM outperforms STM~\cite{wu2025video} by $\mathbf{0.6\%}$ on TGIF-Frame, $\mathbf{0.8\%}$ on MSRVTT-QA, and $\mathbf{0.6\%}$ on ActivityNet-QA, which demonstrates its powerful capability of complex multimodal reasoning.

\begin{table}
  \caption{Video captioning results under CIDEr. The best results are indicated in \textbf{bold}. Models pretrained with massive datasets are noted in \textcolor{gray}{gray}.}
  \label{tab:cap}
  \centering
  \resizebox{0.44\textwidth}{!}{
  \begin{tabular}{cccc}
    \toprule
    Method & \#Pairs & MSRVTT & MSVD \\
    \midrule
    SwinBERT~\cite{lin2022swinbert} & 5M & 53.8 & 120.6 \\
    VIOLETv2~\cite{fu2023empirical} & 5M & 58.0 & 139.2 \\
    LAVENDER~\cite{li2023lavender} & 5M & 58.0 & 139.2 \\
    HiTeA~\cite{ye2023hitea} & 5M & 62.5 & 145.1 \\
    STM~\cite{wu2025video} & 5M & 63.9 & 145.4 \\
    ClusterSTM & 5M & \textbf{64.4} & \textbf{145.6} \\
    \midrule
    \multicolumn{4}{l}{\textcolor{gray}{\emph{Models pretrained on more data}}} \\
    \midrule
    \textcolor{gray}{MV-GPT}~\cite{seo2022end} & \textcolor{gray}{53M} & \textcolor{gray}{60.0} & \textcolor{gray}{-} \\
    \bottomrule
  \end{tabular}
  }
\end{table}

\noindent\textbf{Video Captioning.} Table~\ref{tab:cap} presents the experimental results on the video captioning task. ClusterSTM achieves state-of-the-art performance, with CIDEr scores of $64.4$($\mathbf{+0.5}$) on MSRVTT and $145.6$($\mathbf{+0.2}$) on MSVD, demonstrating its strong capability in video understanding.

\subsection{Ablation Studies}

To evaluate the contribution of each component, we conduct ablation studies on each component. The ablation experiments are trained on a subset composed of half of the WebVid-2M and CC3M datasets, for a total of 5 epochs. We then evaluate the model on the MSRVTT dataset across three downstream tasks. Except for Table~\ref{tab:stm}, we pretrain UMT~\cite{li2023unmasked} for 10 epochs on the full 5M corpus.

\begin{table}
  \caption{Experimental results of different masking strategies on MSRVTT dataset across three downstream tasks. The best results are indicated in \textbf{bold}.}
  \label{tab:ms}
  \centering
  \resizebox{0.45\textwidth}{!}{
  \begin{tabular}{cccc}
    \toprule
    Masking Strategy & Retrieval & QA & Captioning \\
    \midrule
    frame-wise & 33.9 & 34.6 & 50.8 \\
    random & 37.5 & 38.5 & 52.2 \\
    tube-wise & 38.1 & 39.7 & 53.3 \\
    cluster-wise-S & 39.8 & 41.4 & 55.4 \\
    cluster-wise-ST & \textbf{41.3} & \textbf{42.7} & \textbf{56.2} \\
    \bottomrule
  \end{tabular}
  }
\end{table}

\noindent\textbf{Masking Strategies.} Table~\ref{tab:ms} presents the experimental results of different masking strategies on the MSRVTT dataset across three downstream tasks. Notebly, cluster-wise-S denotes randomly selecting one token to retain within each cluster, while cluster-wise-ST denotes retaining the token with the highest temporal density within each cluster-corresponding to our proposed method. Cluster-wise-S performs intra-frame clustering followed by cluster-wise masking, which substantially preserves the complete visual content of video frames. This enables the model to efficiently leverage the retained tokens for representation learning, leading to significant performance gains compared with frame-wise masking, random masking, and tube-wise masking. Furthermore, cluster-wise-ST addresses the issue of temporal information leakage by ensuring strong temporal correlation among the retained tokens, thereby further enhancing the model's ability to comprehend video contextual information.

\begin{table}
  \caption{Experimental results of different masking ratios for image and video tokens on MSRVTT dataset across three downstream tasks. The best results are indicated in \textbf{bold}.}
  \label{tab:mr}
  \centering
  \resizebox{0.45\textwidth}{!}{
  \begin{tabular}{ccccc}
    \toprule
    Image(\%) & Video(\%) & Retrieval & QA & Captioning \\
    \midrule
    50 & 80 & 40.1 & 41.8 & 55.6 \\
    50 & 90 & 41.1 & 42.4 & 56.1 \\
    75 & 80 & 40.9 & 42.1 & 56.0 \\
    75 & 90 & \textbf{41.3} & \textbf{42.7} & \textbf{56.2} \\
    90 & 95 & 39.5 & 41.2 & 55.3 \\
    \bottomrule
  \end{tabular}
  }
\end{table}

\noindent\textbf{Masking Ratios.} We investigate the impact of different masking ratios for image and video tokens, as shown in Table~\ref{tab:mr}. Our model achieves the best performance when using masking ratios of $75\%$ for image tokens and $90\%$ for video tokens. Compared with UMT~\cite{li2023unmasked}(with masking ratios of $50\%$ for image tokens and $80\%$ for video tokens) and STM~\cite{wu2025video}(with a masking ratio of $30\%$ for both image and video tokens), our model benefits from a more balanced masking strategy, allowing it to make more comprehensive use of the data and achieve more efficient pretraining.

\begin{table}
  \caption{Experimental results of different reconstruction targets on MSRVTT dataset across three downstream tasks. The best results are indicated in \textbf{bold}.}
  \label{tab:rt}
  \centering
  \resizebox{0.45\textwidth}{!}{
  \begin{tabular}{cccc}
    \toprule
    Reconstrction Target & Retrieval & QA & Captioning \\
    \midrule
    pixels & 36.3 & 37.4 & 52.1 \\
    HOG & 38.5 & 39.5 & 53.7 \\
    features & 40.2 & 41.5 & 55.3 \\
    video-text relevance & \textbf{41.3} & \textbf{42.7} & \textbf{56.2} \\
    \bottomrule
  \end{tabular}
  }
\end{table}

\noindent\textbf{Reconstruction Targets.} As shown in Figure~\ref{tab:rt}, we compare the experimental results obtained using different reconstruction targets. HOG denotes the Histogram of Oriented Gradients~\cite{dalal2005histograms}, and the features target is supervised by SigLIP~\cite{zhai2023sigmoid}. The model performs the worst when the reconstruction target is low-level pixels. As the semantic level of the reconstruction target increases, the model performance improves. Benefiting from its high-level semantic attributes and multimodal nature, which better align with the cross-modal alignment task, the video-text relevance target enables the model to achieve the best performance.

\begin{table}
  \caption{Effect of different pretraining objectives on MSRVTT dataset across three downstream tasks. The best results are indicated in \textbf{bold}.}
  \label{tab:loss}
  \centering
  \resizebox{0.45\textwidth}{!}{
  \setlength\tabcolsep{3pt}
  \begin{tabular}{ccccccc}
    \toprule
    VTC & VTM & MLM & MRM & Retrieval & QA & Captioning \\
    \midrule
    \checkmark & - & - & - & 36.8 & 37.8 & 51.2 \\
    \checkmark & \checkmark & - & - & 38.2 & 38.9 & 52.7 \\
    \checkmark & \checkmark & \checkmark & - & 38.9 & 39.6 & 54.0 \\
    \checkmark & \checkmark & \checkmark & \checkmark & \textbf{41.3} & \textbf{42.7} & \textbf{56.2} \\
    \bottomrule
  \end{tabular}
  }
\end{table}

\noindent\textbf{Pretraining Objectives.} Table~\ref{tab:loss} presents the experimental results illustrating the impact of different pretraining objectives on model performance. It can be observed that each pretraining objective contributes positively to performance improvement. In particular, MRM yields the most significant performance gain, demonstrating that its capability to align high-level multimodal semantic information greatly enhances the model’s ability for joint visual-textual representation learning.

\begin{table}
  \caption{Effectiveness of cluster-wise spatio-temporal masking strategy. The best results are indicated in \textbf{bold}.}
  \label{tab:stm}
  \centering
  \resizebox{0.45\textwidth}{!}{
  \setlength\tabcolsep{5pt}
  \begin{tabular}{ccccc}
    \toprule
    Method & MSRVTT & DiDeMo & ANet & MSVD \\
    \midrule
    UMT~\cite{li2023unmasked} & 29.6 & 33.4 & 28.3 & 36.2 \\
    + ClusterSTM & \textbf{30.4} & \textbf{34.7} & \textbf{30.2} & \textbf{38.5} \\
    \bottomrule
  \end{tabular}
  }
\end{table}

\noindent\textbf{Effectiveness of Cluster-Wise Spatio-Temporal Masking strategy.} To further validate the effectiveness of the cluster-wise spatio-temporal masking strategy, we replace the semantic masking in UMT~\cite{li2023unmasked} with our proposed cluster-wise spatio-temporal masking while keeping all other model components (with CLIP~\cite{radford2021learning} as the teacher model) unchanged. We report the results under the zero-shot text-to-video retrieval setting, as shown in Figure~\ref{tab:stm}. It can be observed that replacing the masking strategy leads to a significant performance improvement for UMT. Although a high masking ratio causes most visual tokens to be dropped, our masking strategy ensures that the retained tokens comprehensively capture the overall visual content, which is crucial for effective cross-modal alignment.

\section{Conclusion}

In this work, we introduced ClusterSTM, a novel Cluster-Wise Spatio-Temporal Masking strategy tailored for efficient video-language pretraining. By coupling intra-frame clustering with temporal-density-based cluster-wise masking, our approach simultaneously alleviates severe visual information loss and mitigates temporal information leakage—two crucial bottlenecks in masked video modeling. In addition, we proposed a video-text relevance reconstruction objective that enhances high-level multimodal alignment beyond conventional visual reconstruction. Through extensive experiments on multiple video-language benchmarks, ClusterSTM consistently achieved superior performance and strong generalization capability compared to existing approaches. We believe that the proposed cluster-wise spatio-temporal masking paradigm provides a new perspective for efficient multimodal representation learning and can inspire future work toward scalable, temporally coherent video-language foundation models.

\noindent\textbf{Acknowledgments} This work was supported by the National Natural Science Foundation of China (Nos. 62476148, 62501191), and the Guangdong Basic and Applied Basic Research Foundation (No. 2024A1515011292).

%\clearpage
{
    \small
    \bibliographystyle{ieeenat_fullname}
    \bibliography{main}

@String(CVPR= {IEEE Conf. Comput. Vis. Pattern Recog.})

@String(ECCV= {Eur. Conf. Comput. Vis.})

@String(ICLR = {Int. Conf. Learn. Represent.})

@String(AAAI = {AAAI})

@String(CVPR  = {CVPR})

@String(ECCV  = {ECCV})

@String(ICLR  = {ICLR})

@inproceedings{wang2024internvideo2,
  title={Internvideo2: Scaling foundation models for multimodal video understanding},
  author={Wang, Yi and Li, Kunchang and Li, Xinhao and Yu, Jiashuo and He, Yinan and Chen, Guo and Pei, Baoqi and Zheng, Rongkun and Wang, Zun and Shi, Yansong and others},
  booktitle={European Conference on Computer Vision},
  pages={396--416},
  year={2024},
  organization={Springer}
}

@article{bolya2025perception,
  title={Perception encoder: The best visual embeddings are not at the output of the network},
  author={Bolya, Daniel and Huang, Po-Yao and Sun, Peize and Cho, Jang Hyun and Madotto, Andrea and Wei, Chen and Ma, Tengyu and Zhi, Jiale and Rajasegaran, Jathushan and Rasheed, Hanoona and others},
  journal={arXiv preprint arXiv:2504.13181},
  year={2025}
}

@inproceedings{wang2023all,
  title={All in one: Exploring unified video-language pre-training},
  author={Wang, Jinpeng and Ge, Yixiao and Yan, Rui and Ge, Yuying and Lin, Kevin Qinghong and Tsutsui, Satoshi and Lin, Xudong and Cai, Guanyu and Wu, Jianping and Shan, Ying and others},
  booktitle={Proceedings of the IEEE/CVF Conference on Computer Vision and Pattern Recognition},
  pages={6598--6608},
  year={2023}
}

@article{wang2022internvideo,
  title={Internvideo: General video foundation models via generative and discriminative learning},
  author={Wang, Yi and Li, Kunchang and Li, Yizhuo and He, Yinan and Huang, Bingkun and Zhao, Zhiyu and Zhang, Hongjie and Xu, Jilan and Liu, Yi and Wang, Zun and others},
  journal={arXiv preprint arXiv:2212.03191},
  year={2022}
}

@inproceedings{li2023unmasked,
  title={Unmasked teacher: Towards training-efficient video foundation models},
  author={Li, Kunchang and Wang, Yali and Li, Yizhuo and Wang, Yi and He, Yinan and Wang, Limin and Qiao, Yu},
  booktitle={Proceedings of the IEEE/CVF international conference on computer vision},
  pages={19948--19960},
  year={2023}
}

@inproceedings{wu2025video,
  title={Video Language Model Pretraining with Spatio-temporal Masking},
  author={Wu, Yue and Qi, Zhaobo and Sun, Junshu and Wang, Yaowei and Huang, Qingming and Wang, Shuhui},
  booktitle={Proceedings of the Computer Vision and Pattern Recognition Conference},
  pages={8557--8567},
  year={2025}
}

@inproceedings{fu2023empirical,
  title={An empirical study of end-to-end video-language transformers with masked visual modeling},
  author={Fu, Tsu-Jui and Li, Linjie and Gan, Zhe and Lin, Kevin and Wang, William Yang and Wang, Lijuan and Liu, Zicheng},
  booktitle={Proceedings of the IEEE/CVF Conference on Computer Vision and Pattern Recognition},
  pages={22898--22909},
  year={2023}
}

@inproceedings{he2022masked,
  title={Masked autoencoders are scalable vision learners},
  author={He, Kaiming and Chen, Xinlei and Xie, Saining and Li, Yanghao and Doll{\'a}r, Piotr and Girshick, Ross},
  booktitle={Proceedings of the IEEE/CVF conference on computer vision and pattern recognition},
  pages={16000--16009},
  year={2022}
}

@article{tong2022videomae,
  title={Videomae: Masked autoencoders are data-efficient learners for self-supervised video pre-training},
  author={Tong, Zhan and Song, Yibing and Wang, Jue and Wang, Limin},
  journal={Advances in neural information processing systems},
  volume={35},
  pages={10078--10093},
  year={2022}
}

@article{hou2022milan,
  title={Milan: Masked image pretraining on language assisted representation},
  author={Hou, Zejiang and Sun, Fei and Chen, Yen-Kuang and Xie, Yuan and Kung, Sun-Yuan},
  journal={arXiv preprint arXiv:2208.06049},
  year={2022}
}

@inproceedings{ye2023hitea,
  title={Hitea: Hierarchical temporal-aware video-language pre-training},
  author={Ye, Qinghao and Xu, Guohai and Yan, Ming and Xu, Haiyang and Qian, Qi and Zhang, Ji and Huang, Fei},
  booktitle={Proceedings of the IEEE/CVF International Conference on Computer Vision},
  pages={15405--15416},
  year={2023}
}

@inproceedings{wang2024omnivid,
  title={Omnivid: A generative framework for universal video understanding},
  author={Wang, Junke and Chen, Dongdong and Luo, Chong and He, Bo and Yuan, Lu and Wu, Zuxuan and Jiang, Yu-Gang},
  booktitle={Proceedings of the IEEE/CVF conference on computer vision and pattern recognition},
  pages={18209--18220},
  year={2024}
}

@inproceedings{jin2024chat,
  title={Chat-univi: Unified visual representation empowers large language models with image and video understanding},
  author={Jin, Peng and Takanobu, Ryuichi and Zhang, Wancai and Cao, Xiaochun and Yuan, Li},
  booktitle={Proceedings of the IEEE/CVF Conference on Computer Vision and Pattern Recognition},
  pages={13700--13710},
  year={2024}
}

@inproceedings{cheng2023vindlu,
  title={Vindlu: A recipe for effective video-and-language pretraining},
  author={Cheng, Feng and Wang, Xizi and Lei, Jie and Crandall, David and Bansal, Mohit and Bertasius, Gedas},
  booktitle={Proceedings of the IEEE/CVF Conference on Computer Vision and Pattern Recognition},
  pages={10739--10750},
  year={2023}
}

@inproceedings{lei2023revealing,
  title={Revealing single frame bias for video-and-language learning},
  author={Lei, Jie and Berg, Tamara and Bansal, Mohit},
  booktitle={Proceedings of the 61st Annual Meeting of the Association for Computational Linguistics (Volume 1: Long Papers)},
  pages={487--507},
  year={2023}
}

@inproceedings{li2023lavender,
  title={Lavender: Unifying video-language understanding as masked language modeling},
  author={Li, Linjie and Gan, Zhe and Lin, Kevin and Lin, Chung-Ching and Liu, Zicheng and Liu, Ce and Wang, Lijuan},
  booktitle={Proceedings of the IEEE/CVF Conference on Computer Vision and Pattern Recognition},
  pages={23119--23129},
  year={2023}
}

@inproceedings{lin2023smaug,
  title={Smaug: Sparse masked autoencoder for efficient video-language pre-training},
  author={Lin, Yuanze and Wei, Chen and Wang, Huiyu and Yuille, Alan and Xie, Cihang},
  booktitle={Proceedings of the IEEE/CVF International Conference on Computer Vision},
  pages={2459--2469},
  year={2023}
}

@article{bao2021beit,
  title={Beit: Bert pre-training of image transformers},
  author={Bao, Hangbo and Dong, Li and Piao, Songhao and Wei, Furu},
  journal={arXiv preprint arXiv:2106.08254},
  year={2021}
}

@inproceedings{wei2022masked,
  title={Masked feature prediction for self-supervised visual pre-training},
  author={Wei, Chen and Fan, Haoqi and Xie, Saining and Wu, Chao-Yuan and Yuille, Alan and Feichtenhofer, Christoph},
  booktitle={Proceedings of the IEEE/CVF conference on computer vision and pattern recognition},
  pages={14668--14678},
  year={2022}
}

@article{dosovitskiy2020image,
  title={An image is worth 16x16 words: Transformers for image recognition at scale},
  author={Dosovitskiy, Alexey},
  journal={arXiv preprint arXiv:2010.11929},
  year={2020}
}

@inproceedings{devlin2019bert,
  title={Bert: Pre-training of deep bidirectional transformers for language understanding},
  author={Devlin, Jacob and Chang, Ming-Wei and Lee, Kenton and Toutanova, Kristina},
  booktitle={Proceedings of the 2019 conference of the North American chapter of the association for computational linguistics: human language technologies, volume 1 (long and short papers)},
  pages={4171--4186},
  year={2019}
}

@article{rodriguez2014clustering,
  title={Clustering by fast search and find of density peaks},
  author={Rodriguez, Alex and Laio, Alessandro},
  journal={science},
  volume={344},
  number={6191},
  pages={1492--1496},
  year={2014},
  publisher={American Association for the Advancement of Science}
}

@article{shu2022masked,
  title={Masked contrastive pre-training for efficient video-text retrieval},
  author={Shu, Fangxun and Chen, Biaolong and Liao, Yue and Xiao, Shuwen and Sun, Wenyu and Li, Xiaobo and Zhu, Yousong and Wang, Jinqiao and Liu, Si},
  journal={arXiv preprint arXiv:2212.00986},
  year={2022}
}

@inproceedings{zhai2023sigmoid,
  title={Sigmoid loss for language image pre-training},
  author={Zhai, Xiaohua and Mustafa, Basil and Kolesnikov, Alexander and Beyer, Lucas},
  booktitle={Proceedings of the IEEE/CVF international conference on computer vision},
  pages={11975--11986},
  year={2023}
}

@inproceedings{bain2021frozen,
  title={Frozen in time: A joint video and image encoder for end-to-end retrieval},
  author={Bain, Max and Nagrani, Arsha and Varol, G{\"u}l and Zisserman, Andrew},
  booktitle={Proceedings of the IEEE/CVF international conference on computer vision},
  pages={1728--1738},
  year={2021}
}

@inproceedings{sharma2018conceptual,
  title={Conceptual captions: A cleaned, hypernymed, image alt-text dataset for automatic image captioning},
  author={Sharma, Piyush and Ding, Nan and Goodman, Sebastian and Soricut, Radu},
  booktitle={Proceedings of the 56th Annual Meeting of the Association for Computational Linguistics (Volume 1: Long Papers)},
  pages={2556--2565},
  year={2018}
}

@inproceedings{xu2016msr,
  title={Msr-vtt: A large video description dataset for bridging video and language},
  author={Xu, Jun and Mei, Tao and Yao, Ting and Rui, Yong},
  booktitle={Proceedings of the IEEE conference on computer vision and pattern recognition},
  pages={5288--5296},
  year={2016}
}

@inproceedings{anne2017localizing,
  title={Localizing moments in video with natural language},
  author={Anne Hendricks, Lisa and Wang, Oliver and Shechtman, Eli and Sivic, Josef and Darrell, Trevor and Russell, Bryan},
  booktitle={Proceedings of the IEEE international conference on computer vision},
  pages={5803--5812},
  year={2017}
}

@inproceedings{chen2011collecting,
  title={Collecting highly parallel data for paraphrase evaluation},
  author={Chen, David and Dolan, William B},
  booktitle={Proceedings of the 49th annual meeting of the association for computational linguistics: human language technologies},
  pages={190--200},
  year={2011}
}

@inproceedings{krishna2017dense,
  title={Dense-captioning events in videos},
  author={Krishna, Ranjay and Hata, Kenji and Ren, Frederic and Fei-Fei, Li and Carlos Niebles, Juan},
  booktitle={Proceedings of the IEEE international conference on computer vision},
  pages={706--715},
  year={2017}
}

@inproceedings{jang2017tgif,
  title={Tgif-qa: Toward spatio-temporal reasoning in visual question answering},
  author={Jang, Yunseok and Song, Yale and Yu, Youngjae and Kim, Youngjin and Kim, Gunhee},
  booktitle={Proceedings of the IEEE conference on computer vision and pattern recognition},
  pages={2758--2766},
  year={2017}
}

@inproceedings{yu2018joint,
  title={A joint sequence fusion model for video question answering and retrieval},
  author={Yu, Youngjae and Kim, Jongseok and Kim, Gunhee},
  booktitle={Proceedings of the European conference on computer vision (ECCV)},
  pages={471--487},
  year={2018}
}

@inproceedings{xu2017video,
  title={Video question answering via gradually refined attention over appearance and motion},
  author={Xu, Dejing and Zhao, Zhou and Xiao, Jun and Wu, Fei and Zhang, Hanwang and He, Xiangnan and Zhuang, Yueting},
  booktitle={Proceedings of the 25th ACM international conference on Multimedia},
  pages={1645--1653},
  year={2017}
}

@inproceedings{yu2019activitynet,
  title={Activitynet-qa: A dataset for understanding complex web videos via question answering},
  author={Yu, Zhou and Xu, Dejing and Yu, Jun and Yu, Ting and Zhao, Zhou and Zhuang, Yueting and Tao, Dacheng},
  booktitle={Proceedings of the AAAI Conference on Artificial Intelligence},
  volume={33},
  number={01},
  pages={9127--9134},
  year={2019}
}

@article{li2022uniformerv2,
  title={Uniformerv2: Spatiotemporal learning by arming image vits with video uniformer},
  author={Li, Kunchang and Wang, Yali and He, Yinan and Li, Yizhuo and Wang, Yi and Wang, Limin and Qiao, Yu},
  journal={arXiv preprint arXiv:2211.09552},
  year={2022}
}

@article{loshchilov2017fixing,
  title={Fixing weight decay regularization in adam},
  author={Loshchilov, Ilya and Hutter, Frank and others},
  journal={arXiv preprint arXiv:1711.05101},
  volume={5},
  number={5},
  pages={5},
  year={2017}
}

@article{fu2021violet,
  title={Violet: End-to-end video-language transformers with masked visual-token modeling},
  author={Fu, Tsu-Jui and Li, Linjie and Gan, Zhe and Lin, Kevin and Wang, William Yang and Wang, Lijuan and Liu, Zicheng},
  journal={arXiv preprint arXiv:2111.12681},
  year={2021}
}

@article{wang2022omnivl,
  title={Omnivl: One foundation model for image-language and video-language tasks},
  author={Wang, Junke and Chen, Dongdong and Wu, Zuxuan and Luo, Chong and Zhou, Luowei and Zhao, Yucheng and Xie, Yujia and Liu, Ce and Jiang, Yu-Gang and Yuan, Lu},
  journal={Advances in neural information processing systems},
  volume={35},
  pages={5696--5710},
  year={2022}
}

@article{luo2022clip4clip,
  title={Clip4clip: An empirical study of clip for end to end video clip retrieval and captioning},
  author={Luo, Huaishao and Ji, Lei and Zhong, Ming and Chen, Yang and Lei, Wen and Duan, Nan and Li, Tianrui},
  journal={Neurocomputing},
  volume={508},
  pages={293--304},
  year={2022},
  publisher={Elsevier}
}

@inproceedings{lei2021less,
  title={Less is more: Clipbert for video-and-language learning via sparse sampling},
  author={Lei, Jie and Li, Linjie and Zhou, Luowei and Gan, Zhe and Berg, Tamara L and Bansal, Mohit and Liu, Jingjing},
  booktitle={Proceedings of the IEEE/CVF conference on computer vision and pattern recognition},
  pages={7331--7341},
  year={2021}
}

@inproceedings{li2022align,
  title={Align and prompt: Video-and-language pre-training with entity prompts},
  author={Li, Dongxu and Li, Junnan and Li, Hongdong and Niebles, Juan Carlos and Hoi, Steven CH},
  booktitle={Proceedings of the IEEE/CVF conference on computer vision and pattern recognition},
  pages={4953--4963},
  year={2022}
}

@inproceedings{ge2022bridging,
  title={Bridging video-text retrieval with multiple choice questions},
  author={Ge, Yuying and Ge, Yixiao and Liu, Xihui and Li, Dian and Shan, Ying and Qie, Xiaohu and Luo, Ping},
  booktitle={Proceedings of the IEEE/CVF conference on computer vision and pattern recognition},
  pages={16167--16176},
  year={2022}
}

@inproceedings{wang2023unified,
  title={Unified coarse-to-fine alignment for video-text retrieval},
  author={Wang, Ziyang and Sung, Yi-Lin and Cheng, Feng and Bertasius, Gedas and Bansal, Mohit},
  booktitle={Proceedings of the IEEE/CVF international conference on computer vision},
  pages={2816--2827},
  year={2023}
}

@article{chen2023vast,
  title={Vast: A vision-audio-subtitle-text omni-modality foundation model and dataset},
  author={Chen, Sihan and Li, Handong and Wang, Qunbo and Zhao, Zijia and Sun, Mingzhen and Zhu, Xinxin and Liu, Jing},
  journal={Advances in Neural Information Processing Systems},
  volume={36},
  pages={72842--72866},
  year={2023}
}

@inproceedings{yang2021just,
  title={Just ask: Learning to answer questions from millions of narrated videos},
  author={Yang, Antoine and Miech, Antoine and Sivic, Josef and Laptev, Ivan and Schmid, Cordelia},
  booktitle={Proceedings of the IEEE/CVF international conference on computer vision},
  pages={1686--1697},
  year={2021}
}

@article{zellers2021merlot,
  title={Merlot: Multimodal neural script knowledge models},
  author={Zellers, Rowan and Lu, Ximing and Hessel, Jack and Yu, Youngjae and Park, Jae Sung and Cao, Jize and Farhadi, Ali and Choi, Yejin},
  journal={Advances in neural information processing systems},
  volume={34},
  pages={23634--23651},
  year={2021}
}

@inproceedings{seo2022end,
  title={End-to-end generative pretraining for multimodal video captioning},
  author={Seo, Paul Hongsuck and Nagrani, Arsha and Arnab, Anurag and Schmid, Cordelia},
  booktitle={Proceedings of the IEEE/CVF conference on computer vision and pattern recognition},
  pages={17959--17968},
  year={2022}
}

@inproceedings{lin2022swinbert,
  title={Swinbert: End-to-end transformers with sparse attention for video captioning},
  author={Lin, Kevin and Li, Linjie and Lin, Chung-Ching and Ahmed, Faisal and Gan, Zhe and Liu, Zicheng and Lu, Yumao and Wang, Lijuan},
  booktitle={Proceedings of the IEEE/CVF conference on computer vision and pattern recognition},
  pages={17949--17958},
  year={2022}
}

@inproceedings{ma2022x,
  title={X-clip: End-to-end multi-grained contrastive learning for video-text retrieval},
  author={Ma, Yiwei and Xu, Guohai and Sun, Xiaoshuai and Yan, Ming and Zhang, Ji and Ji, Rongrong},
  booktitle={Proceedings of the 30th ACM international conference on multimedia},
  pages={638--647},
  year={2022}
}

@inproceedings{xiao2021next,
  title={Next-qa: Next phase of question-answering to explaining temporal actions},
  author={Xiao, Junbin and Shang, Xindi and Yao, Angela and Chua, Tat-Seng},
  booktitle={Proceedings of the IEEE/CVF conference on computer vision and pattern recognition},
  pages={9777--9786},
  year={2021}
}

@inproceedings{yang2023vid2seq,
  title={Vid2seq: Large-scale pretraining of a visual language model for dense video captioning},
  author={Yang, Antoine and Nagrani, Arsha and Seo, Paul Hongsuck and Miech, Antoine and Pont-Tuset, Jordi and Laptev, Ivan and Sivic, Josef and Schmid, Cordelia},
  booktitle={Proceedings of the IEEE/CVF conference on computer vision and pattern recognition},
  pages={10714--10726},
  year={2023}
}

@article{zhuang2025spatial,
  title={Spatial-temporal saliency guided unbiased contrastive learning for video scene graph generation},
  author={Zhuang, Weijun and Dong, Bowen and Zhu, Zhilin and Li, Zhijun and Liu, Jie and Wang, Yaowei and Hong, Xiaopeng and Li, Xin and Zuo, Wangmeng},
  journal={IEEE Transactions on Multimedia},
  year={2025},
  publisher={IEEE}
}

@inproceedings{dalal2005histograms,
  title={Histograms of oriented gradients for human detection},
  author={Dalal, Navneet and Triggs, Bill},
  booktitle={2005 IEEE computer society conference on computer vision and pattern recognition (CVPR'05)},
  volume={1},
  pages={886--893},
  year={2005},
  organization={Ieee}
}

@inproceedings{radford2021learning,
  title={Learning transferable visual models from natural language supervision},
  author={Radford, Alec and Kim, Jong Wook and Hallacy, Chris and Ramesh, Aditya and Goh, Gabriel and Agarwal, Sandhini and Sastry, Girish and Askell, Amanda and Mishkin, Pamela and Clark, Jack and others},
  booktitle={International conference on machine learning},
  pages={8748--8763},
  year={2021},
  organization={PmLR}
}

@inproceedings{li2023blip,
  title={Blip-2: Bootstrapping language-image pre-training with frozen image encoders and large language models},
  author={Li, Junnan and Li, Dongxu and Savarese, Silvio and Hoi, Steven},
  booktitle={International conference on machine learning},
  pages={19730--19742},
  year={2023},
  organization={PMLR}
}

@inproceedings{baevski2022data2vec,
  title={Data2vec: A general framework for self-supervised learning in speech, vision and language},
  author={Baevski, Alexei and Hsu, Wei-Ning and Xu, Qiantong and Babu, Arun and Gu, Jiatao and Auli, Michael},
  booktitle={International conference on machine learning},
  pages={1298--1312},
  year={2022},
  organization={PMLR}
}

@article{zhou2021ibot,
  title={ibot: Image bert pre-training with online tokenizer},
  author={Zhou, Jinghao and Wei, Chen and Wang, Huiyu and Shen, Wei and Xie, Cihang and Yuille, Alan and Kong, Tao},
  journal={arXiv preprint arXiv:2111.07832},
  year={2021}
}

@inproceedings{liu2023devil,
  title={The devil is in the frequency: Geminated gestalt autoencoder for self-supervised visual pre-training},
  author={Liu, Hao and Jiang, Xinghua and Li, Xin and Guo, Antai and Hu, Yiqing and Jiang, Deqiang and Ren, Bo},
  booktitle={Proceedings of the AAAI Conference on Artificial Intelligence},
  volume={37},
  number={2},
  pages={1649--1656},
  year={2023}
}

@article{xie2022masked,
  title={Masked frequency modeling for self-supervised visual pre-training},
  author={Xie, Jiahao and Li, Wei and Zhan, Xiaohang and Liu, Ziwei and Ong, Yew Soon and Loy, Chen Change},
  journal={arXiv preprint arXiv:2206.07706},
  year={2022}
}

@article{feichtenhofer2022masked,
  title={Masked autoencoders as spatiotemporal learners},
  author={Feichtenhofer, Christoph and Li, Yanghao and He, Kaiming and others},
  journal={Advances in neural information processing systems},
  volume={35},
  pages={35946--35958},
  year={2022}
}

@inproceedings{wang2022bevt,
  title={Bevt: Bert pretraining of video transformers},
  author={Wang, Rui and Chen, Dongdong and Wu, Zuxuan and Chen, Yinpeng and Dai, Xiyang and Liu, Mengchen and Jiang, Yu-Gang and Zhou, Luowei and Yuan, Lu},
  booktitle={Proceedings of the IEEE/CVF conference on computer vision and pattern recognition},
  pages={14733--14743},
  year={2022}
}

@inproceedings{li2023citetracker,
  title={Citetracker: Correlating image and text for visual tracking},
  author={Li, Xin and Huang, Yuqing and He, Zhenyu and Wang, Yaowei and Lu, Huchuan and Yang, Ming-Hsuan},
  booktitle={Proceedings of the IEEE/CVF international conference on computer vision},
  pages={9974--9983},
  year={2023}
}

@inproceedings{huang2024rtracker,
  title={Rtracker: Recoverable tracking via pn tree structured memory},
  author={Huang, Yuqing and Li, Xin and Zhou, Zikun and Wang, Yaowei and He, Zhenyu and Yang, Ming-Hsuan},
  booktitle={Proceedings of the IEEE/CVF Conference on Computer Vision and Pattern Recognition},
  pages={19038--19047},
  year={2024}
}

@inproceedings{MengLLJ025,
  author       = {Weikang Meng and
                  Yadan Luo and
                  Xin Li and
                  Dongmei Jiang and
                  Zheng Zhang},
  title        = {PolaFormer: Polarity-aware Linear Attention for Vision Transformers},
  booktitle    = {The Thirteenth International Conference on Learning Representations,
                  {ICLR} 2025, Singapore, April 24-28, 2025},
  year         = {2025}
}

@article{meng2025nalaformer,
  title={Nalaformer: Norm-aware linear attention for transformer models},
  author={Meng, Weikang and Luo, Yadan and Huo, Liangyu and Wang, Yaowei and Li, Xin and Zhang, Zheng},
  journal={arXiv e-prints},
  pages={arXiv--2506},
  year={2025}
}
}


\end{document}

\setcounter{page}{1}
\maketitlesupplementary

%##########################################################
\begin{strip}
  \centering
  \includegraphics[width=0.98\linewidth]{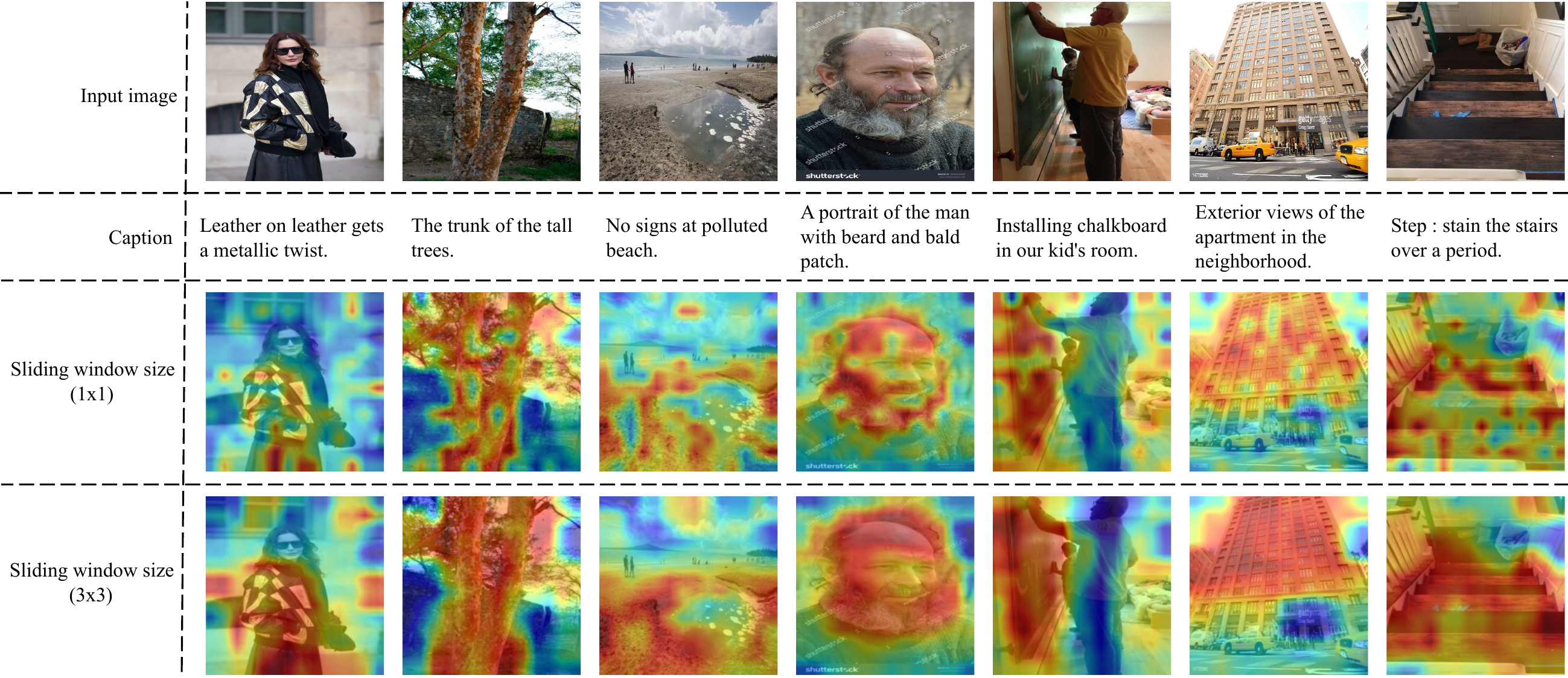}
  \captionof{figure}{Visualization of the relevance heatmaps generated with different sliding window size. From top to bottom, we show the input image, the corresponding caption, the relevance heatmap obtained with a 1x1 sliding window, and the relevance heatmap obtained with a 3x3 sliding window.}
  \label{fig:vis_rel}
\end{strip}
%##########################################################

\section{Additional Analysis}

\noindent\textbf{Different Sliding Window Size for Relevance Generation.} To investigate the effect of different sliding window sizes on the quality of video-text relevance generation, we generate relevance matrices using sliding windows of size 1x1 and 3x3, respectively. The visualization of the relevance heatmaps are shown in Figure~\ref{fig:vis_rel}. As illustrated, directly interacting a single visual token with the text feature fails to produce high-quality relevance scores that accurately capture the correspondence between visual and textual information. This is primarily because the teacher model is trained with global vision-language contrastive learning, which leads it to pay limited attention to local semantic details. In contrast, after aggregating neighboring tokens, the resulting relevance maps exhibit significantly improved quality.

Meanwhile, Table~\ref{tab:sw} reports the experimental results obtained with different sliding window sizes across various downstream tasks. As shown, using a 3x3 sliding window leads to substantial performance gains over the 1x1 setting, further demonstrating the effectiveness of our relevance generation strategy.

%##########################################################
\begin{table}[htb]
  \caption{Experimental results of different sliding window sizes for relevance generation on MSRVTT~\cite{xu2016msr} dataset across three downstream tasks. The best results are indicated in \textbf{bold}.}
  \label{tab:sw}
  \centering
  %\resizebox{0.47\textwidth}{!}{
  %\setlength\tabcolsep{5pt}
  \begin{tabular}{cccc}
    \toprule
    Sliding Window Size & Retrieval & QA & Captioning \\
    \midrule
    1x1 & 39.7 & 40.2 & 54.7 \\
    3x3 & \textbf{41.3} & \textbf{42.7} & \textbf{56.2} \\
    \bottomrule
  \end{tabular}
  %}
\end{table}
%##########################################################

\begin{figure*}[htb]
  \centering
  \includegraphics[width=0.9\linewidth]{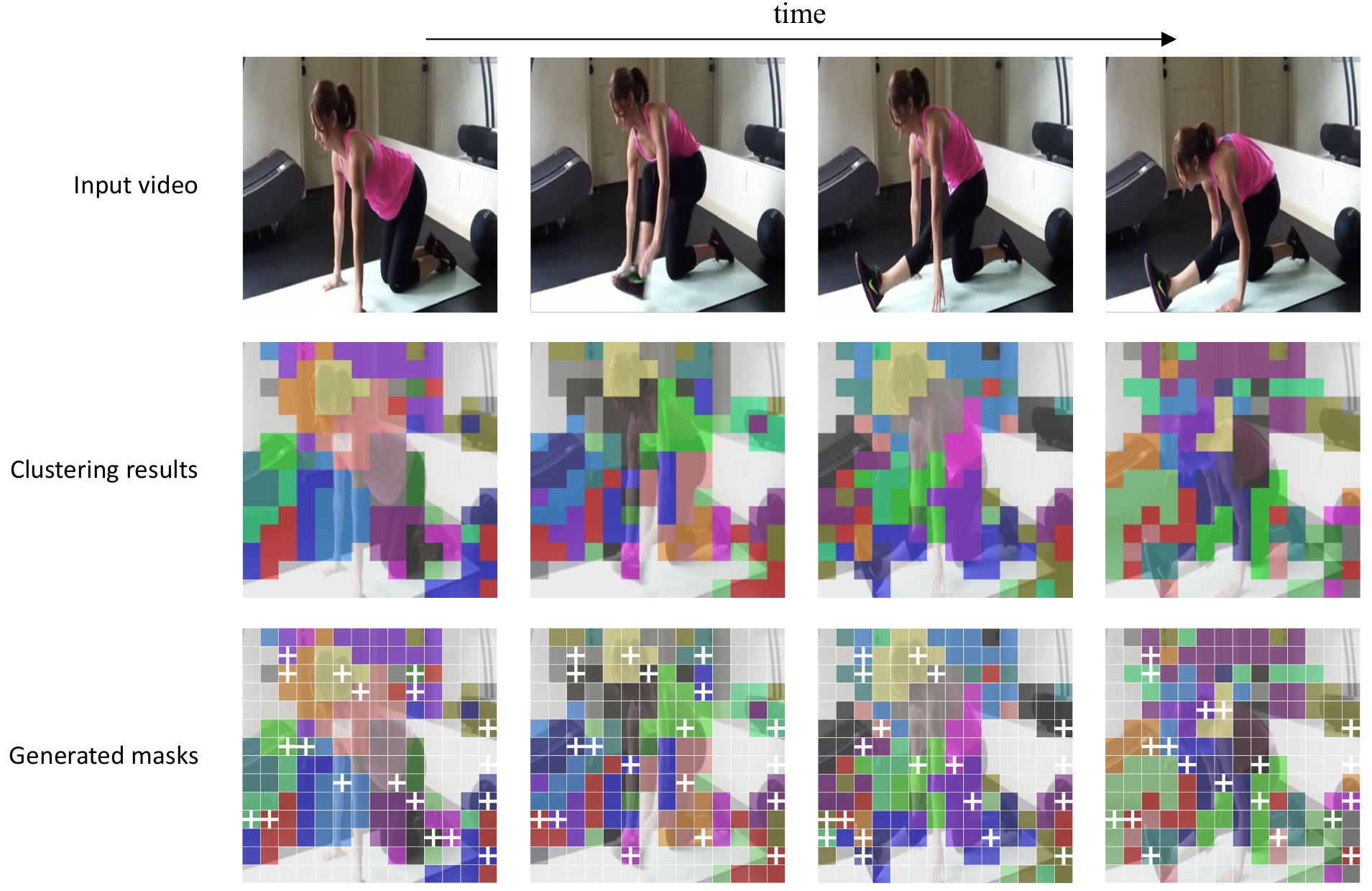}
  \caption{Visualization of the cluster-wise spatio-temporal masking. From top to bottom, we show the input video, the clustering results, and the generated masks, where "+" indicates the tokens retained by the masking strategy.}
  \label{fig:vis_masks}
\end{figure*}

\noindent\textbf{Qualitative Results of Cluster-wise Spatio-Temporal Masks.} As shown in Figure~\ref{fig:vis_masks}, we present the visualization of the masks generated by our cluster-wise spatio-temporal masking strategy. The cluster-wise masking ensures that the retained tokens collectively preserve the complete visual content of the video. Moreover, our masking strategy successfully captures the dynamic motion cues of the \textit{head} and \textit{legs} during the \textit{yoga} action, and, guided by temporal density, retains the tokens that are most semantically correlated over time. As a result, the preserved tokens exhibit strong temporal consistency.

%##########################################################
%##########################################################

\noindent\textbf{Limitations and Future Work.} Although cluster-wise spatio-temporal masking demonstrates promising effectiveness in addressing the severe visual information loss and temporal information leakage encountered when applying masked visual modeling to large-scale video-language pretraining, it relies heavily on the quality of clustering, and the clustering algorithm we adopt is non-learnable, which constrains its adaptability to diverse video domains and dynamic spatio-temporal patterns. A potential direction for future work is to design learnable spatio-temporal deep clustering algorithms that can be jointly optimized with the pretraining objectives. Such approaches could enable the clustering process to better align with high-level video-language semantics and dynamically adapt to scene variations.

\section{More Implementation Details}

\noindent\textbf{Pseudocode of Cluster-Wise Spatio-Temporal Masking.} Intra-frame clustering divides visual tokens into multiple semantically independent clusters, and the number of tokens within each cluster often varies. This irregularity poses challenges for designing code that can efficiently process multiple video inputs in parallel. Therefore, we provide PyTorch-style pseudocode for parallel cluster-wise spatio-temporal masking, as shown in Algorithms~\ref{alg:clusterstm} and~\ref{alg:tdc}.

%##########################################################
\begin{algorithm}[t]
  \caption{PyTorch-style pseudocode for parallel cluster-wise spatio-temporal masking.}
  \label{alg:clusterstm}
  \definecolor{codegreen}{rgb}{0.25,0.75,0.25}
  \definecolor{codered}{rgb}{0.75,0.25,0.25}
  \lstset{
    backgroundcolor=\color{white},
    basicstyle=\fontsize{7pt}{7pt}\ttfamily\selectfont,
    columns=fullflexible,
    breaklines=true,
    captionpos=b,
    commentstyle=\fontsize{7pt}{7pt}\color{codegreen},
    stringstyle=\fontsize{7pt}{7pt}\color{codered},
    keywordstyle=\fontsize{7pt}{7pt},
    %frame=tb,
  }
  \begin{lstlisting}[language=python]
def ClusterSTM(x, mask_ratio, dc_ratio):
  # x: normalized visual tokens
  # mask_ratio: masking ratio applied for masking
  # dc_ratio: quantile percentage for cutoff distance
  
  B,T,N,C = x.shape
  num_clusters = N * (1 - mask_ratio)
  
  #------- Step 1: Density Peak Clusteing (DPC) -------
  clusters = DPC(x, num_clusters)
  clusters = clusters.reshape(B*T,N)
  
  #------- Step 2: Temporal Density Calculation -------
  temp_rho = TDC(x, dc_ratio)
  
  #----------- Step 3: Cluster-Wise Masking -----------
  flat_rho = temp_rho.reshape(-1)
  flat_clusters = clusters.reshape(-1)
  
  # add offsets to cluster IDs
  max_clusters = clusters.max(dim=1).values+1
  clusters_offset = torch.cat([torch.tensor([0]), max_clusters.cumsum(0)[:-1]])
  global_offset = clusters_offset.repeat_interleave(N)
  global_clusters = flat_clusters + global_offset
  
  # cluster-wise softmax
  num_global_clusters = clusters_offset[-1] + max_clusters[-1]
  sum_rho = torch.zeros(num_global_clusters).scatter_add_(0, global_clusters, flat_rho)
  global_scores = fllat_rho / (sum_rho[global_clusters] + 1e-12)
  
  # cluster-wise argmax
  max_per_cluster = torch.full((num_global_clusters,), -1e9)
  max_per_cluster.scatter_reduce_(0, global_clusters, global_scores, reduce="amax", include_self=True)
  
  # masking
  masks = global_scores != max_per_cluster[global_clusters]
  masks = masks.reshape(B,T,N)
  return masks
  \end{lstlisting}
\end{algorithm}
%##########################################################

%##########################################################
\begin{algorithm}[t]
  \caption{PyTorch-style pseudocode for parallel temporal density calculation.}
  \label{alg:tdc}
  \definecolor{codegreen}{rgb}{0.25,0.75,0.25}
  \definecolor{codered}{rgb}{0.75,0.25,0.25}
  \lstset{
    backgroundcolor=\color{white},
    basicstyle=\fontsize{7pt}{7pt}\ttfamily\selectfont,
    columns=fullflexible,
    breaklines=true,
    captionpos=b,
    commentstyle=\fontsize{7pt}{7pt}\color{codegreen},
    stringstyle=\fontsize{7pt}{7pt}\color{codered},
    keywordstyle=\fontsize{7pt}{7pt},
    %frame=tb,
  }
  \begin{lstlisting}[language=python]
def TDC(x, dc_ratio):
  # x: normalized visual tokens
  # dc_ratio: quantile percentage for cutoff distance
  
  B,T,N,C = x.shape
  
  spat_x = x.reshape(B*T,N,C)
  temp_x = x.unsqueeze(1).expand(-1,T,-1,-1,-1)
  mask = ~torch.eye(T)
  temp_x = temp_x[:,mask].reshape(B*T,(T-1)*N,C)

  # compute the temporal semantic distances between each token and all tokens in neighboring frames
  temp_dist = 1 - torch.matmul(spat_x, temp_x.transpose(-1,-2))

  # compute the cutoff distance
  dc = torch.quantile(temp_dist.view(B*T,-1), dc_ratio, dim=1, keepdim=True).view(B*T,1,1)

  # compute temporal density based on the temporal semantic distances
  temp_rho = torch.sum(torch.exp(-(temp_dist / dc)**2), dim=-1) / N
  return temp_rho
  \end{lstlisting}
\end{algorithm}
%##########################################################

\noindent\textbf{Pseudocode of Video-Text Relevance Generation.} The usage of sliding windows poses challenges for performing relevance generation in parallel. To address this, we provide PyTorch-style pseudocode for parallel relevance generation, as shown in Algorithm~\ref{alg:rg}.

%##########################################################
\begin{algorithm}[t]
  \caption{PyTorch-style pseudocode for parallel relevance generation.}
  \label{alg:rg}
  \definecolor{codegreen}{rgb}{0.25,0.75,0.25}
  \definecolor{codered}{rgb}{0.75,0.25,0.25}
  \lstset{
    backgroundcolor=\color{white},
    basicstyle=\fontsize{7pt}{7pt}\ttfamily\selectfont,
    columns=fullflexible,
    breaklines=true,
    captionpos=b,
    commentstyle=\fontsize{7pt}{7pt}\color{codegreen},
    stringstyle=\fontsize{7pt}{7pt}\color{codered},
    keywordstyle=\fontsize{7pt}{7pt},
    %frame=tb,
  }
  \begin{lstlisting}[language=python]
def RG(x, y, Pooler):
  # x: visual tokens output by the vision encoder
  # y: text feature output by the text encoder
  # Pooler: pooling module of the teacher model
  
  B,T,N,C = x.shape
  L = int(N**0.5)
  
  x = x.permute(0,1,3,2).reshape(B*T,C,L,L) # B*T,C,L,L
  x = torch.nn.functional.pad(x, (1,1,1,1), mode="replicate") # B*T,C,L+2,L+2
  x = x.unflod(2,3,1).unfload(3,3,1) # B*T,C,L,L,3,3
  x = x.permute(0,2,3,4,5,1) # B*T,L,L,3,3,C
  x = x.reshape(-1,9,C) # B*T*L*L,9,C
  x = Pooler(x) # B*T*L*L,C
  x = x.reshape(B,T*N,C) # B,T*N,C
  
  x_norm = x / x.norm(p=2, dim=-1, keepdim=True) # B,T*N,C
  y_norm = y / y.norm(p=2, dim=-1, keepdim=True) # B,C
  relevance = torch.einsum("bnc,bc->bn", x_norm, y_norm) # B,T*N
  relevance = relevance.reshape(B,T,N)
  return relevance
  \end{lstlisting}
\end{algorithm}
%##########################################################

{
    \small
    \bibliographystyle{ieeenat_fullname}
    \bibliography{main}
}
